\documentclass[journal,comsoc]{IEEEtran}
\usepackage[T1]{fontenc}
\usepackage{cite}

\ifCLASSINFOpdf
   \usepackage[pdftex]{graphicx}
\else
   \usepackage[dvips]{graphicx}
\fi

\usepackage{amsmath}
\usepackage[cmintegrals]{newtxmath}

\usepackage[numbers]{natbib}
\usepackage{booktabs}
\usepackage{multirow}
\usepackage{multicol}
\usepackage{amsmath}
\usepackage{algpseudocode}
\usepackage{mathtools}
\usepackage[linesnumbered,ruled,vlined]{algorithm2e}

\mathchardef\mhyphen="2D

\begin{document}
%
\title{Human Simulation Computation: A Human-Inspired Framework for Adaptive AI Systems}

\author{Hong~Su
\IEEEcompsocitemizethanks{\IEEEcompsocthanksitem H. Su is with the School of Computer Science, Chengdu University of Information Technology, Chengdu, China.\\
 E-mail: suguest@126.com. \\
\protect\\
}
\thanks{}}

\markboth{Journal of \LaTeX\ Class Files,~Vol.~14, No.~8, August~2015}%
{Shell \MakeLowercase{\textit{et al.}}: Bare Demo of IEEEtran.cls for IEEE Communications Society Journals}
%

\maketitle

\begin{abstract}
Large language models (LLMs) have demonstrated strong capabilities in knowledge representation and reasoning based on textual data. However, their reliance on language material alone limits their ability to adapt, verify reasoning outcomes, and operate effectively in open and dynamic real-world environments. In this paper, we propose Human Simulation Computation (HSC), a human-inspired computational framework that models intelligence as a continuous, closed-loop process involving thinking, action, learning, reflection, and activity scheduling, collectively referred to as the internal reasoning process. HSC emphasizes active participation both within the internal reasoning process and in interactions with the environment, where actions are used not only to achieve goals but also to automatically refine and improve internal reasoning mechanisms without external intervention. Furthermore, HSC incorporates commonly used human thinking strategies across all stages of the internal reasoning process, such as main-feature-oriented reasoning, scope expansion through action, and on-time learning driven by environmental feedback. Through theoretical analysis, we argue that human simulation strategies cannot be fully learned from language material alone, and that human-like reasoning processes and action-grounded reasoning methods are essential for robust adaptation and effective interaction with real-world environments.
\end{abstract}

\begin{IEEEkeywords}
    Human Simulation Computation; Environment Interaction; Adaptive Artificial Intelligence; Human-Inspired Reasoning
\end{IEEEkeywords}

\IEEEpeerreviewmaketitle

\section{Introduction}

Humans interact with the world through an integrated process of thinking,
goal formation, action execution, learning, and reflection.
This process enables individuals to adapt to complex and changing environments,
rather than merely solving isolated, predefined tasks.
The environment itself is not limited to the physical world; it also includes
other intelligent agents, artificial systems, and even language-based models
that participate in interaction and feedback.

Recent advances in large language models (LLMs) \cite{chang2024survey} have demonstrated impressive
capabilities in language understanding, reasoning, and knowledge generation.
However, LLMs are fundamentally trained on static language corpora.
As a result, they do not possess intrinsic mechanisms for goal-driven behavior,
environmental interaction, or action-based verification.
Their reasoning remains largely detached from the real world, relying on
linguistic consistency rather than grounded feedback.
Consequently, LLMs alone cannot fully reproduce the adaptive processes that
humans use to live and operate effectively in dynamic environments.

This observation raises a fundamental question:  
\emph{Can we simulate the way humans live and adapt—through thinking, learning,
action, reflection, and feedback—so that AI systems can improve through direct
interaction with their environments rather than through language alone?}
Moreover, unlike task-oriented systems designed for narrow objectives, both
humans and intelligent agents must continuously adapt to new situations,
uncertainties, and long-term goals.
An AI system that aims to operate robustly in real environments \cite{su2026actively} must therefore
go beyond task execution and develop the capability to ``live better'' through
ongoing interaction and self-improvement.

To address this challenge, we introduce \emph{Human Simulation Computation}
(HSC), a human-inspired framework that models intelligence as a systematic,
interactive, and continuous process.
HSC consists of two tightly coupled aspects.
The first is a simulation procedure that integrates thinking, learning,
reflection, action, and feedback acquisition into a closed loop.
In this loop, thinking guides decision making; learning records outcomes \cite{su2025human} and
internal processes in an on-time manner; reflection evaluates and updates all
stages; and actions actively influence both the internal cognitive process and
the external environment.
Unlike passive observation, actions in HSC are used to accelerate feedback,
verify reasoning, and promote learning, including the learning of action
sequences that reduce future computational cost.

The second aspect of HSC is the incorporation of commonly used human thinking
strategies throughout the simulation process.
Prior work has shown that introducing specific reasoning patterns, such as
chain-of-thought prompting, can improve LLM performance on individual problems.
However, such approaches are typically problem-specific and lack a unified,
systematic structure.
In contrast, HSC embeds human-inspired thinking strategies—such as difference
detection, scope expansion, candidate comparison, and reflective reasoning—into
each stage of the simulation loop, enabling consistent adaptation across tasks
and environments.

A key feature of HSC is that learning is not restricted to external language
data.
Instead, the system learns from everything it processes or senses, including
thinking paths, action sequences, reflection outcomes, and environmental
feedback over time.
Actions play a particularly important role: they not only affect the external
world, but also modify internal processes by promoting learning, triggering
reflection, and validating reasoning through grounded feedback.
In this sense, actions are not predefined task operators, but mechanisms for
achieving better long-term adaptation to the environment.

\textbf{Contributions.}
The main contributions of this paper are summarized as follows:
\begin{itemize}
    \item We propose Human Simulation Computation, a framework that simulates
    the full human interaction process with the environment—including thinking,
    learning, reflection, action, and feedback—with the objective of long-term
    adaptation rather than solving isolated predefined tasks.

    \item We identify and formalize commonly used human thinking strategies and
    integrate them into each stage of the human simulation process, enabling AI
    systems to exhibit human-like adaptive behavior in early stages of
    deployment.

    \item We emphasize the systematic interaction among thinking, learning,
    reflection, and action, and introduce an activity control mechanism that
    schedules these processes, including background learning and reflection
    during idle periods.

    \item We demonstrate the importance of action as an active driver of
    learning and reasoning, enabling on-time data acquisition, internal process
    optimization, and environment-grounded verification without reliance on
    external intervention.
\end{itemize}

The remainder of this paper is organized as follows.
Section~II reviews related work.
Section~III introduces the overall framework of Human Simulation Computation for AI systems.
Section~IV presents human simulation thinking as the cognitive foundation of the proposed model.
Section~V describes the human simulation activity scheduling mechanism.
Section~VI discusses human simulation reflection and retrospection.
Section~VII elaborates on human simulation learning.
Section~VIII explains action mechanisms for interacting with the environment and other AI systems.
Section~IX provides theoretical analysis and proofs.
Finally, we conclude the paper.

\section{Related Work}

This work relates to several research directions, including large language
models, reasoning-enhanced prompting methods, embodied and interactive AI,
and human-inspired cognitive architectures.
We review these areas and clarify how Human Simulation Computation (HSC)
differs from and extends existing approaches.

\subsection{Large Language Models}

Large language models (LLMs) have achieved remarkable success in natural
language understanding, generation, and reasoning by learning from large-scale
text corpora~\cite{chang2024survey, naveed2025comprehensive}.
Recent models demonstrate strong performance on a wide range of tasks, including
question answering, code generation \cite{vaithilingam2022expectation}, and mathematical reasoning.

However, LLMs are primarily trained on static language data and operate by
predicting token sequences conditioned on textual context.
As a result, LLMs lack intrinsic mechanisms for goal formation, environment
interaction, and action-based verification.
Their reasoning processes are evaluated mainly through linguistic consistency
rather than grounded feedback from the real world~\cite{novikova2025consistency}.
Although LLMs can describe actions or simulate interaction symbolically, they
do not directly experience the consequences of those actions.

\subsection{Reasoning-Enhanced Prompting and Chain-of-Thought}

To improve reasoning capability, several studies have introduced prompting
strategies such as chain-of-thought (CoT) \cite{lee2024applying}, self-consistency, and reflection-based
reasoning~\cite{cheng2025vision}.
These methods guide LLMs to generate intermediate reasoning steps, which can
improve performance on complex tasks.

While effective, these approaches remain fundamentally language-driven.
They focus on improving reasoning within a single inference episode or a single
problem instance.
The reasoning process is not systematically linked to long-term learning,
environmental interaction, or action outcomes.
In contrast, HSC embeds reasoning strategies into a continuous loop that
includes action execution, feedback acquisition, reflection, and on-time
learning, enabling persistent adaptation beyond isolated problem solving.

\subsection{Embodied and Interactive Artificial Intelligence}

Embodied AI and interactive learning emphasize the importance of grounding
intelligence in perception and action~\cite{memarian2024embodied, zhao2024embodied}.
By interacting with physical or simulated environments, agents can acquire
experience that cannot be obtained from language data alone.
Reinforcement learning and embodied navigation are representative examples of
this paradigm~\cite{li2020unsupervised, bigazzi2021out}.

However, many embodied AI systems rely on predefined reward functions, fixed
task definitions, or narrowly scoped environments.
Their learning objectives are typically task-centric rather than oriented toward
long-term adaptive living.
Moreover, the internal reasoning and reflection processes of such agents are
often implicit or opaque.
HSC differs by explicitly modeling thinking, learning, reflection, and action as
separate but interacting stages, with actions serving not only as task operators
but also as tools for verification and self-improvement.

\subsection{Cognitive Architectures and Human-Inspired Models}

Cognitive architectures aim to model human cognition by integrating perception,
memory, reasoning, and action~\cite{sumers2023cognitive}.
These systems emphasize structured internal processes and long-term knowledge
accumulation.
Human-inspired reasoning strategies, such as heuristic search and
difference-reduction methods, have been studied in classical artificial
intelligence~\cite{addison2024human}.

HSC builds upon this tradition but differs in two important aspects.
First, it explicitly incorporates modern LLMs as components within a broader
human simulation framework rather than treating symbolic reasoning as the sole
intelligence mechanism.
Second, HSC emphasizes continuous interaction with open environments and
on-time learning from both internal processes and external feedback, rather than
operating within closed or predefined cognitive models.

\subsection{Summary and Positioning}

In summary, existing approaches either focus on language-based reasoning without
grounded interaction, or on environment interaction without explicit modeling of
human-like thinking and reflection.
Human Simulation Computation bridges this gap by unifying thinking, action,
learning, reflection, and scheduling into a systematic, interactive framework.
By treating adaptation and ``living better'' as primary objectives, HSC provides
a complementary perspective to both LLM-centric reasoning methods and
task-oriented embodied AI systems.

\section{Human Simulation Computation of AI System}

Humans are shaped by natural selection; consequently, a fundamental capability
of a person is to adapt to and live within natural environments.
Following this principle, an AI system designed for real-world deployment
should also be capable of adapting to its environment.

Various factors trigger an AI system to perform actions.
These factors are referred to as \textbf{action trigger sources}.
Action trigger sources include the internal goals of the AI system as well as
external environmental feedback, including feedback from other AI systems.

Large Language Models (LLMs) are trained on large-scale human language data.
As a result, they do not directly learn human thinking processes, but instead
learn linguistic patterns that are correlated with thinking.
Language is a product of thinking, and a single thinking mode may generate
language expressions across many scenarios.
Therefore, learning language alone does not imply the acquisition of general
thinking methods.
To enable adaptive intelligence, explicit thinking modes need to be provided
to LLM-based systems.

\textbf{Scope limitation:}
In this paper, we focus on human simulation computation that can be realized
with the assistance of LLMs and programmable systems.
Cognitive abilities that are difficult to formalize computationally, such as
imagination, are beyond the scope of this work.

\subsection{Model}

The human-inspired model aims to enable an AI system to live and adapt in a
complex environment, rather than relying on predefined actions.
The system learns, thinks, and takes actions as an active participant in its
environment.
Accordingly, the design objective of the AI system is to improve its ability to
operate within the environment, similar to how humans adapt over time.

Based on this perspective, we propose \emph{Human Simulation Computation} (HSC),
also referred to as human-inspired computation.
Inspired by human thinking processes, an AI system under HSC actively interacts
with the environment and other AI systems.
The AI system applies its own thinking to decide when and how to act, with the
goal of adapting to the environment rather than completing isolated tasks.
Unlike conventional systems that function as passive observers, the proposed
agent operates as an active participant, capable of triggering, delaying, or
modifying environmental events through deliberate actions.

Formally, the HSC process can be described as a closed-loop state transition, as
shown in Eq.~(\ref{eq:hsc_state}).
The internal cognitive state $s_t$ is first processed by a thinking function
$\mathcal{T}(\cdot)$, conditioned on environmental factors $f_t$.
Based on this cognitive assessment, an action is selected via
$\mathcal{A}(\cdot)$.
The outcomes of the action are then evaluated through a reflection mechanism
$\mathcal{R}(\cdot)$, and the internal state is subsequently updated by a
learning function $\mathcal{L}(\cdot)$, yielding the next state $s_{t+1}$.

\begin{equation}
\label{eq:hsc_state}
s_{t+1} = \mathcal{L}\!\left(
    s_t,\;
    \mathcal{R}\!\left(
        s_t,\;
        \mathcal{A}\!\left(
            s_t,\;
            \mathcal{T}(s_t, f_t)
        \right)
    \right)
\right)
\end{equation}

To adapt to the environment effectively, humans typically perform a recurring
process of thinking, learning, action, and reflection.
This process, referred to as the \emph{human simulation stages}, forms a
continuous interaction with the environment.
Human Simulation Computation follows the same principle and exhibits the
following characteristics:

(1) \textbf{Thinking-driven decision making:}
Thinking determines whether an action should be taken, how it should be
executed, and when it should occur.
For example, the system may decide to reject learning outcomes that are harmful
to itself.

(2) \textbf{Active participation through action:}
Actions are taken to actively participate in the environment rather than
remaining a passive observer.
Action selection is guided by thinking, including decisions on whether to act,
how to act, and whether to reuse existing methods.
Actions are also used to acquire information for internal reasoning process (thinking, reflection, learning et.al.). Such as obtaining additional data or accelerating environmental feedback.
Instead of waiting for events to occur naturally, the system may act
proactively to shorten feedback cycles, as discussed in~\cite{su2026actively}.

(3) \textbf{Reflection (Reflectionaction):}
Reflection is used to identify problems and opportunities for improvement
across all stages.
After actions are executed, the system evaluates whether the outcomes are
effective and may reflect on prior thinking, actions, and even the reflection
process itself.

(4) \textbf{Continuous learning:}
Learning accumulates experience from all stages of the human simulation
process.
Not only outcomes, but also thinking patterns, action strategies, and
reflection results are learned and retained over time.

(5) \textbf{Information acquisition from multiple sources:}
Information is obtained from the environment, other AI systems, and the system
itself.
All acquired information is examined through thinking processes to assess
correctness, completeness, and consistency.
The environment provides diverse contexts and challenges, which offer broader
training opportunities and guide the evolution of thinking strategies.

In conventional systems, environmental factors directly lead to results,
without intermediate cognitive intervention:
\begin{equation}
\label{eq_common_process}
\text{factors} \Rightarrow \text{results}.
\end{equation}

In contrast, Human Simulation Computation introduces intermediate stages of
thinking, action, reflection, and learning between factors and results.
Actions are repeatedly interleaved with cognitive processing, forming a
progressive decision process:
\begin{equation}
\label{eq_hun_process}
\left.\begin{matrix}
    \left.\begin{matrix} 
        \left.\begin{matrix} 
            \text{factors} \\ 
            \text{action}
        \end{matrix}\right\}\Rightarrow & m_1 \\
                                        & \text{action}
    \end{matrix}\right\}\Rightarrow & m_2 \\
                                    & \text{action}
\end{matrix}\right\}\Rightarrow m_3 \Rightarrow \text{results}.
\end{equation}

This formulation emphasizes that results are not produced solely by external
factors, but emerge through multiple rounds of cognition and interaction.
Each intermediate stage may influence subsequent actions and reasoning,
allowing the system to adapt dynamically rather than responding in a single
step.

\subsubsection{Using Human Thinking Modes to Guide LLM Reasoning}

Human thinking modes are explicitly provided to the LLM as structured inputs.
Since an LLM operates as a conditional prediction model, injecting human
simulation thinking methods can guide the LLM toward corresponding reasoning
patterns.
In this way, the LLM is encouraged to follow human-like thinking processes
rather than producing responses based solely on surface-level language
patterns.

\subsection{Systematic, Interactive, and On-Time Operation}

The human simulation process is systematic and forms a closed-loop structure.
Learning and action mutually reinforce each other: learning guides future
actions, while actions are deliberately designed to enhance learning.
For example, when a new method is identified, the system may perform additional
actions to generate more data, thereby reinforcing memory and improving
retention.
Through this interaction, the AI system can operate autonomously without
continuous external intervention.

In the following sections, we further analyze the human simulation computation
process from two perspectives: adopting human thinking modes that can evolve
with experience, and enabling actions that participate actively in the current
situation rather than relying on predefined responses.

\subsubsection{On-Time Processing}

Human simulation processes must operate in an on-time manner to adapt to
changing environments.
Thinking should respond promptly to sudden changes, as future conditions
cannot be fully anticipated and solutions may need to be adjusted dynamically.
This behavior mirrors human adaptation to evolving situations.

Learning should also be performed on time by incorporating newly sensed
information into the LLM or explicit recording mechanisms.
Action sequences learned over time and space, which can only be acquired
through real-world interaction, should be updated continuously.

Actions must likewise be executed in a timely manner.
Delayed actions may become ineffective if environmental conditions change,
highlighting the importance of on-time thinking, learning, and action for
successful adaptation.

\section{Human Simulation Thinking}

Human simulation thinking is a core component of Human Simulation Computation.
It aims to simulate human thinking processes in order to identify ways for an
AI system to live and operate more effectively in complex environments, based
on encountered problems and environmental characteristics.
Thinking is responsible for analyzing situations and determining whether to
act, how to act, and when to act.
It governs both internal cognitive processes and interactions with the external
environment, such as rejecting learning outcomes that may be harmful to the
system itself.

The thinking process is not predefined.
Instead, it dynamically evaluates the environment and identifies the main
questions or key features relevant to the current situation.
Thinking guides the entire human simulation process by generating solutions
and recommending actions at each stage.

The AI system adopts human-like thinking modes to guide its interaction with
the environment.
As a result, the same stimulus may lead to different outcomes depending on the
system’s accumulated experience and internal judgment.
For example, a red color may be interpreted as dangerous by a system with prior
experience related to injury, while in other contexts it may be considered
neutral.
Similarly, actions such as moving slowly or quickly are not inherently good or
bad, but are determined by situational assessment and learned experience.

\subsection{Goal-Oriented Thinking}

Thinking processes are inherently directional and are guided by the goals of the
AI system.
This design is inspired by human behavior, where actions are not performed for
isolated tasks, but for continuous self-improvement and adaptation within the
environment.
Accordingly, the primary objective of the AI system is to maintain and enhance
its ability to exist and operate effectively in dynamic surroundings.
When optimal performance cannot be achieved, goal-oriented thinking shifts
toward minimizing harm or maximizing benefit, either at the physical or
cognitive level.

Thinking does not directly request the LLM to generate a solution for a
specific task.
Instead, the AI system provides goal-oriented constraints and preferences,
which are applied either before LLM generation or during the selection of
generated candidates.
Through this process, the system chooses solutions that best align with its
current goals.

Goals may differ in scope and priority.
They may focus on task completion, self-benefit, or benefit to others.
The AI system may also assign different levels of emphasis to certain goals,
reflecting their relative importance.
Such emphasis can be incorporated into LLM prompts to guide generation toward
preferred outcomes.

\subsubsection{Learnable Goals}

Goals are not static; they evolve through experience and reflection.
A goal represents a high-level principle that can be applied across multiple
tasks and methods.
For example, if an autonomous vehicle initially prioritizes speed and
subsequently experiences accidents, reflection may identify excessive speed as
a major contributing factor.
As a result, the system updates its goal hierarchy to place greater emphasis on
safety.

Separating goals from task-specific methods enables reuse across different
contexts.
A goal such as safety can be applied to driving, work processes, or interactions
with other AI systems.
If such principles were embedded independently into each detailed task, the
resulting learning process would become redundant and inefficient.
By decoupling goals from specific methods, the AI system achieves improved
flexibility and scalability.

\subsubsection{Trigger Sources}

Human Simulation Computation is not strictly task-driven; therefore, specific
factors are required to trigger the AI system to initiate thinking or action.
These factors are referred to as \emph{trigger sources}.

Goals serve as primary trigger sources by driving the AI system to perform
systematic actions in pursuit of desired outcomes.
The overarching goal of living better within the environment is a fundamental
trigger source.
This goal can be further divided into basic maintenance requirements, such as
preserving system power and avoiding physical or functional damage, which
ensure the system’s continued operation.

Beyond basic survival, higher-level trigger sources focus on improving quality
of operation.
These include maintaining stable internal states, responding to feedback from
other AI systems (e.g., avoiding actions that cause harm to others), and
considering environmental impact, such as pollution or interference.

Some goals operate over long time horizons and cannot be satisfied through
immediate feedback.
Such goals require long-term planning and delayed evaluation.
Examples include tolerating temporary discomfort for long-term benefit, such as
exercise-induced pain or the acquired tolerance of bitter tastes.

\subsection{Common Human Thinking Strategies}

Human thinking employs a variety of intuitive strategies to manage complex and dynamic environments. These strategies are grounded in the way people naturally interact with their surroundings. They can guide AI systems to concentrate on pertinent information, broaden reasoning when needed, and adjust decisions based on accumulated experience.  
The following subsections outline several representative thinking strategies incorporated into Human Simulation Computation.

\subsubsection{Main-Feature-Oriented Reasoning}

Real-world environments involve numerous factors, many of which may be
irrelevant to the problem under consideration.
Accordingly, the thinking process focuses on identifying the main features or
core problems that distinguish the current situation.
This focus guides the LLM to concentrate on essential information rather than
being distracted by unrelated factors.

The main feature or question is typically characterized by differences from
expected or normal conditions.
Such differences can be identified by comparing observed outcomes with expected
results, historical behavior, feedback from other agents, internal states, or
progress toward goals~\cite{su2025difference}.

When the detected difference exceeds a certain threshold, the situation is
considered \emph{abnormal}.
In such cases, the system should expand its reasoning scope.
As an illustration, consider a visual cue such as "1/9/18." If the slashes are uniform in length—deviating from conventional fractional notation—the system must avoid an immediate fractional interpretation. The appropriate strategy is to actively query for broader contextual information to resolve the ambiguity before committing to an inference.

\subsubsection{Wider-Scope Consideration}

Effective thinking requires consideration of a sufficiently broad scope.
To achieve reliable results, the AI system must evaluate whether the current
reasoning scope is adequate.
If relevant information is missing, the system should explicitly recognize this
limitation and take actions to acquire additional data.

Scope expansion may involve spatial, temporal, or relational dimensions.
For example, when deciding the acceleration speed of an autonomous vehicle, the
system should consider not only traffic conditions but also the presence of
passengers.
Such contextual factors cannot always be predefined and must be dynamically
verified through interaction with the environment.

Expanding the reasoning scope often requires action.
The AI system may actively obtain information from a wider spatial area, over a
longer time horizon, or from additional sources, thereby improving the quality
of subsequent decisions.

\subsubsection{Continuous Thinking Before Action}

Continuous thinking refers to repeatedly evaluating a situation over time in
order to compare alternative methods or incorporate newly acquired
information.
The notion of ``thinking multiple times before action'' reflects a common human
practice aimed at improving decision quality by considering additional options
and broader context.

In Human Simulation Computation, continuous thinking enables the system to revise decisions as new information emerges or to select more effective methods based on the evolving context. In urgent scenarios, this may involve delivering an initial prompt response, which is then progressively refined as additional information is obtained.

For example, when approaching a red light on a multi-lane road, the immediate
decision may be to stop.
If subsequent information indicates that adjacent lanes are also congested,
the system may determine that changing lanes is unnecessary.
This illustrates how decisions evolve through continuous thinking as more
information is acquired.

\subsubsection{Generalization and Degeneralization}

Generalization aims to extract more general methods from specific experiences so
that they can be applied across different contexts.
Such generalization may occur across space, time, or different AI systems.
For example, if a method is effective in one location, the system may attempt to
apply it in another.
Similarly, a method learned in the past may be reused in future situations, or
a strategy effective for one agent may be transferred to others.

Degeneralization addresses the opposite scenario.
When a generalized method proves ineffective or inappropriate for a specific
case, the system identifies the distinguishing features of that case and adapts
the method accordingly.
This process requires recognizing new categories or conditions and modifying
existing methods to fit them.
Both generalization and degeneralization rely on prompting the LLM when
inconsistencies or failures are detected.

\subsubsection{Oppositional and Holistic Thinking}

Human thinking often considers problems from opposing perspectives or extends
local observations to a broader context.
These thinking modes help prevent premature conclusions and uncover alternative
solutions.

Oppositional thinking \cite{nolan2025cultivating} involves evaluating both a proposition and its opposite.
For example, when considering whether an AI system should proceed along a wet
road, the system may also evaluate the alternative of not proceeding and assess
the consequences of both options.
Such comparison supports more balanced decision making.

Holistic thinking extends reasoning from a single point or partial observation
to the entire process or system.
If a problem is observed in one part of a spatial or temporal sequence, similar
issues may exist elsewhere.
This mode of reasoning enables the system to infer system-level patterns from
localized evidence~\cite{su2025scatter}.

\subsubsection{Risk-Avoidance and Positive Reframing}

In many situations, humans proactively consider potential negative outcomes in
order to avoid serious damage.
Accordingly, the AI system should assess possible risks in advance and take
preventive actions when the consequences may be severe.
This risk-avoidance thinking focuses on identifying what could go wrong and
reducing the likelihood or impact of failure.

Conversely, when repeated negative outcomes occur and cannot be easily changed,
humans often adopt positive reframing to maintain stability.
For example, the loss of social connections may initially be perceived as
negative, but may also be interpreted as an opportunity to identify more
meaningful relationships.
Similarly, an AI system can learn to reframe persistent failures in a constructive
manner, supporting long-term resilience.

Risk-avoidance and positive reframing function as complementary thinking modes,
encouraging the system to consider both adverse and favorable aspects of a
situation.

\subsubsection{Planning Multiple Candidate Actions}

Planning multiple candidate actions and anticipating their possible outcomes
provides the AI system with greater flexibility in decision making.
By comparing alternative actions in advance, the system can better evaluate
potential benefits and risks before execution.

This strategy is particularly useful for preventing irreversible failures.
By considering contingency plans and rescue options, the system can avoid
committing to actions that offer no opportunity for recovery.
As a result, planning multiple candidate actions supports safer and more robust
behavior in dynamic environments.

\subsection{Action as a Participant in Thinking}

In Human Simulation Computation, action is not merely an execution step but an
active participant in the thinking process.
Actions are used not only to change the external environment, but also to
support and refine internal reasoning.
By interacting with the environment, actions generate feedback that informs
subsequent thinking, learning, and reflection.

Through this interaction, thinking and action form a tightly coupled process.
Thinking guides action selection, while actions provide new information that
may confirm, challenge, or revise existing assumptions.
This bidirectional relationship enables the AI system to adapt its reasoning
based on real-world outcomes rather than relying solely on internal inference.

\subsubsection{Scope Expansion Through Action}

As discussed in the human thinking strategies above, effective reasoning often
requires a sufficiently broad scope.
When the current scope is limited, actions are needed to expand it, such as
actively acquiring information from a wider spatial or temporal range.

In real-world environments, sensing capabilities are often constrained, and it
is neither feasible nor efficient to obtain complete information in advance.
Additionally, other AI systems may operate without sharing direct physical
representations of the environment.
As a result, thinking based on limited scope may lead to suboptimal decisions.

For example, when only vehicle dynamics are considered in autonomous driving,
high speed may appear acceptable.
However, when passengers are present, additional factors such as comfort and
safety must be considered.
Because such contextual factors are dynamic and difficult to predefine, the
thinking process should continuously verify whether its current scope is
sufficient and take actions to expand it when necessary.

\subsubsection{Identifying Hidden Objects and Context}

When attempting to solve a problem, it is important to identify not only the
observable information, but also the hidden objects and contextual factors
related to the target system.
Often, only partial information is explicitly presented, while critical
background details remain implicit.

For example, when diagnosing issues related to accessing a self-deployed LLM,
a system may attempt to answer the question solely based on the query content.
However, if multiple LLM instances are deployed on the same machine using the
same default port, requests may be routed incorrectly, leading to unexpected
behavior.
Without considering the deployment context, such issues cannot be effectively resolved, as users often provide only the information pertinent to the question itself.

Therefore, the AI system should be guided to identify the underlying target
system and the relevant hidden context behind observable symptoms.
This includes recognizing related objects and acquiring additional information
when necessary.
Before generating a response, the system should verify whether sufficient
context has been obtained and actively request further information if
uncertainty remains.
Maintaining a cautious attitude toward incomplete information and taking
actions to obtain clarification are essential aspects of human-like reasoning.

\subsubsection{Evolution of Thinking Modes via Learning and Reflection}

Although thinking modes guide all other processes, they are not fixed.
Through learning and reflection, thinking modes themselves can be updated and
refined over time.
New thinking strategies may be accumulated based on experience, either through
fine-tuning mechanisms or explicit recording methods~\cite{su2025human}.

In addition, actions that expand the reasoning scope can reveal hidden factors
that influence the thinking process.
By incorporating both explicit feedback and newly discovered contextual
information, the AI system can gradually improve its thinking modes and adapt
them to more complex or previously unseen situations.

\subsubsection{Reward for Itself}

The reward of an AI agent may include predefined evaluation metrics; however, it
should ultimately be \emph{aligned with the agent’s goals and regulated by its
internal thinking process, which may evolve over time by on-time learning}.

Reward assignment is jointly determined by goals and internal reasoning.
An action may yield short-term positive outcomes, such as external rewards or
immediate satisfaction; however, if it conflicts with the long-term objective
of living better in the environment (for example, by causing self-damage), the
agent may reinterpret such outcomes as penalties rather than rewards.
This mechanism is inspired by human behavior, where activities such as drinking
or smoking may provide temporary pleasure but are ultimately recognized as
harmful and therefore discouraged.

Rewards can also be learned through experience.
When actions are repeatedly observed to produce unfavorable long-term effects,
the agent updates its internal reward model accordingly, enabling more
appropriate decision-making in future interactions.

\section{Human Simulation Activity Schedule}

The human simulation activity schedule is designed to trigger and organize the
Human Simulation Computation process.
In human behavior, individuals continuously think, recall past experiences to
identify strengths and weaknesses, and learn periodically during free time.
Similarly, an AI system requires a mechanism to schedule these activities,
referred to as \emph{activity control}, which invokes thinking, recall, and
reflection processes.

During idle or low-load periods, the system reviews past events and outcomes,
using explicit features to form associations, resolve unresolved questions, or
derive improved methods.
This process does not aim to revisit all past experiences indiscriminately.
Instead, the schedule prioritizes: (1) issues that are most important to the AI
system, (2) issues that are relatively new, and (3) issues whose outcomes
deviate from normal expectations, with the goal of identifying problems,
improving methods, or avoiding negative results.

Human simulation activity scheduling differs from trigger sources.
Trigger sources determine \emph{why} the AI system initiates actions, whereas
the activity schedule determines \emph{how} and \emph{when} specific thinking,
learning, and reflection tasks are organized and executed.

In the following, we introduce commonly used thinking strategies and actions
employed within the human simulation activity schedule.

\subsection{Most Important Issues Schedule}

Since the overarching goal of the AI system is to live and operate better within
its environment, issue importance is defined relative to this objective.
Certain issues are inherently critical to the system’s basic operation, such
as fundamental maintenance requirements.
These include ensuring sufficient power supply, avoiding physical or functional
damage, and detecting abnormal sensor readings.
Such issues are treated with the highest scheduling priority.

Other important issues arise from feedback provided by the environment or by
other AI systems.
Because the AI system interacts with external entities, it must consider the
effects of its actions on others and respond appropriately to external feedback.
This behavior mirrors human social awareness, where individuals prioritize
feedback from others as part of adaptive decision making.

\subsection{Max-Entropy-Based Schedule}

The activity schedule prioritizes issues that differ most from normal patterns
or that carry the highest informational value.
This mechanism simulates the way humans naturally focus on unusual or
salient events.

The first step is to identify differences.
Differences may appear across time, space, visual perception, sound, or other
sensor measurements, as well as within learned knowledge or expected results.
Temporal differences include unusual timing or frequency of events, while
spatial differences involve occurrences concentrated in specific locations or
distributed across multiple locations.
Visual differences may arise from color, shape, or symbolic patterns, and
auditory differences from abnormal sounds.
Discrepancies between expected and observed outcomes, or between past
experience and current results, are also treated as differences that warrant
further processing.

Importantly, judgments may evolve over time.
An event that appears insignificant in the short term may reveal greater
importance when evaluated over a longer period or in light of accumulated
experience.
Accordingly, the system may re-evaluate past observations during idle periods,
allowing long-term patterns to emerge through reflection and learning.

\subsection{Unsolved Questions}

Issues that remain unresolved or still raise questions are scheduled for further
analysis, particularly during idle periods.
Such issues represent differences at the thinking level and indicate gaps in
current understanding.

The system first attempts to reuse known methods to address unsolved questions.
This is achieved by identifying similar past problems and evaluating whether
previously effective solutions can be transferred.
Similarity-based iteration allows the system to efficiently explore potential
solutions before resorting to entirely new approaches.

\subsubsection{Reason Finding}
\label{sec_reason_found}

Reason finding aims to identify the root causes of unresolved or newly observed
issues.
For example, when a product fails intermittently, the system should determine
whether the failure should be tolerated, mitigated, or eliminated.
Such failures may originate from incomplete design assumptions or limited
experience at the time of development.

By identifying root causes, the system can provide meaningful explanations to
the LLM and support informed decisions about subsequent actions.
This process helps improve methods, refine thinking strategies, and guide
future planning.

\subsection{New Issues Schedule}

When an issue has not been encountered before, it is treated as a new issue.
Humans naturally pay attention to novelty, and this behavior is incorporated
into Human Simulation Computation.
New issues are therefore given higher scheduling priority, as they may provide
opportunities for learning and the generation of new ideas.

An issue does not remain new indefinitely.
As experience accumulates, previously novel situations may become common.
The degree of novelty can be measured by similarity to past experiences.
When similarity exceeds a predefined threshold, the issue is treated as common;
conversely, issues with low similarity or unusually large deviations are
considered significant differences and scheduled for further analysis.

\subsubsection{Creation During Spare Time}

When routine tasks have been completed, the system may select unresolved human
problems or previously difficult issues for further exploration during idle
periods.
Instead of applying known methods, the system may attempt alternative
approaches, encouraging creative problem solving.

Engaging with new issues contributes to experience growth.
During spare time, the system may actively seek novel problems or ideas, while
still evaluating whether such exploration is beneficial.
Since novelty does not always imply usefulness, thinking processes are applied
to determine whether newly encountered issues are worth learning.

\subsection{Actions to Participate in Scheduling}

Actions are also required to support and refine the activity scheduling process.
Through action, the AI system can monitor its operational state and allocate
computational resources to appropriate thinking, learning, or reflection tasks.
Scheduling actions enable the system to balance ongoing task execution with
background cognitive activities.
\subsubsection{Scheduling Tasks During Free Time}

This module monitors the current utilization of system resources, such as CPU
and memory.
When resource usage remains low for a relatively long period—serving as the
equivalent of human free time—the system schedules background activities for
execution.

Scheduling priority is determined by a combination of issue importance,
max-entropy measures, and the presence of new issues.
This strategy reflects human cognitive behavior, where high-priority or unusual
matters are more likely to be considered during idle periods.
\subsubsection{Learning to Schedule}

Scheduling behavior itself can be learned.
Past scheduling decisions and their outcomes are provided to the LLM and used
to improve future scheduling strategies.
In addition to learned patterns, a periodic random trigger is retained to
invoke scheduling, ensuring exploration beyond habitual behavior.

Scheduling factors may also evolve over time.
The system initially considers known factors, such as importance, novelty, and
entropy, and learns how to combine them effectively.
At the same time, previously unknown factors may be discovered through
experience.
By observing contextual conditions and feedback outcomes, the system learns to
schedule activities even when explicit factors are not predefined.
\subsubsection{Suppressing Long-Unsolved Results}

If certain tasks repeatedly receive high priority but remain unsolved over an
extended period, their scheduling priority may be gradually reduced.
This prevents the system from expending excessive resources on issues that do
not yield meaningful progress.

Similarly, differences that persist for a long time without leading to
significant outcomes may be treated as lower priority.
Conversely, seemingly minor differences may be promoted if experience or
prediction suggests that they could lead to important consequences.
Through this adaptive adjustment, the scheduling process remains efficient and
responsive to both short-term feedback and long-term trends.

\section{Human Simulation Reflection}

Humans typically reflect after completing an activity in order to identify
problems or determine whether better methods exist \cite{wan2025srpo}.
An AI system can incorporate a similar reflection process.
All stages of Human Simulation Computation—including thinking, action, learning,
and reflection itself—can be Reflectionspectively examined to identify potential
issues or opportunities for improvement.
The Reflection process should explicitly consider feedback from the environment.

During Reflection, the system recalls past events over time and reviews both the
overall process and its details in relation to observed outcomes.
The recall process should include relevant agents, locations, and procedures as
comprehensively as possible.
Reflection is first conducted at a holistic level to evaluate the entire
process, and then at a detailed level in sequence to refine specific steps.
This examination aims to determine whether problems exist or whether improved
methods can be derived.
Related objects, including other AI systems and environmental factors, are also
taken into account.

\subsection{Commonly Used Ways}

During Reflection, the thinking process considers the entire activity sequence along
the timeline, together with feedback obtained from the environment and other
AI systems.
The whole process is reviewed by examining interactions among all related
objects, including environmental conditions and other agents.

Common reflective questions inspired by human thinking include:
\begin{itemize}
    \item Are there unresolved problems?
    \item Could the process be performed better?
    \item Can problem--method pairs be extracted?
    \item Can the derived methods be applied at other times or to other agents?
\end{itemize}

Additional reflective strategies include summarizing experiences, extracting
general rules, and identifying new or rare patterns with high entropy.

\subsubsection{Increase Thinking Scope}

During Reflection, the thinking process should consider a wider range of factors;
otherwise, important causes may be overlooked due to incomplete information.
Reflection should therefore expand across longer time spans, broader spatial
contexts, and additional related agents in order to identify deeper root
factors.

Rather than focusing solely on factors observed in the current situation, the
system should also examine influences that may emerge over extended periods,
across wider environments, or through interactions with other AI systems.
This broader scope helps uncover hidden dependencies and delayed effects.

When a method is identified as effective or ineffective, its impact should also
be evaluated across a wider scope.
For example, a method that works well in one location may or may not be
applicable in a broader context~\cite{su2025improving}.
Similarly, considering alternative or opposite conditions can reveal additional
insights, such as evaluating both the benefits of performing an action and the
consequences of not performing it.

\subsubsection{Reason Finding}

Reason finding during Reflection focuses on identifying the underlying causes of
observed outcomes in order to improve future thinking, actions, and learning
processes.
This process aims to explain why certain results occurred and how corresponding
methods or decisions can be refined.

This form of reason finding is related to, but distinct from, the process
described in Section~\ref{sec_reason_found}.
While Section~\ref{sec_reason_found} emphasizes discovering reasons for new or
unresolved issues during free time, Reflection-oriented reason finding focuses on
explaining completed processes and their outcomes.
By identifying root causes Reflectionspectively, the system can adjust its methods,
thinking strategies, and action policies more effectively.

\subsubsection{Non-Obvious Association Discovery}

Not all meaningful relationships in the real world are direct cause--effect
pairs.
Some associations are not immediately obvious and may emerge only through
co-occurrence in time or space.
Identifying such non-obvious associations enables the AI system to develop a
richer understanding of environmental patterns.

For example, the presence of blood on the floor may not directly indicate a
specific outcome, but it is often associated with injury or danger.
Although this relationship is not a strict causal link, recognizing the
association supports more informed reasoning and action.
Discovering such patterns often requires active participation and contextual
interpretation rather than passive observation.

Non-obvious association discovery typically involves two steps.
First, the system identifies related features that appear together, such as an
observable event and a later outcome.
Second, it associates these elements through reasoning, even when no direct
causal chain is immediately evident.
Virtual associations may also be formed through internal simulation, such as
mentally evaluating potential responses before they are executed in the real
environment.

\textbf{Indirect Association.}
Many existing rules in AI systems focus on direct cause--effect relationships.
In contrast, indirect associations arise through higher-level reasoning.
For instance, repeated exposure to danger may lead an agent to retreat, but the
agent may also choose to remain based on prior experience or higher-level goals.
Such decisions are not direct reactions, but outcomes of accumulated reasoning
and reflection.

\subsection{Actions to Participate in Reflection}

Actions can also be used to actively support and improve the Reflection process.
Rather than relying solely on passive recall, the AI system may perform actions
that facilitate more effective reflection.
These actions help refine thinking, validate conclusions, and strengthen the
cause--result relationships identified during Reflection.
\subsubsection{Using Thinking to Guide the Reflection Process}

During Reflection, thinking plays a guiding role in structuring and prioritizing the
reflection process.
Previously defined thinking strategies can be reused to analyze past actions,
learning outcomes, and decisions.
Reflection should also be grounded in actual outcomes, using real environmental
feedback as the primary reference.

The Reflection process may itself become the object of reflection.
Thinking, learning, and action strategies applied in earlier stages can all be
reviewed Reflectionspectively.
Together, these processes form a closed loop of cause and result, enabling
continuous self-improvement.
\subsubsection{On-Time Learning to Record Reflection Results and Processes}

The results of the Reflection process should be learned and recorded on time, as they
directly contribute to improving future thinking, action, and learning
processes.
Timely recording ensures that valuable insights obtained through reflection are
not lost and can be reused when similar situations arise.

In addition to recording Reflection outcomes, the Reflection methods themselves should
also be learned.
Some reflection strategies may prove more effective than others as experience
accumulates.
By learning which Reflection approaches lead to better improvements, the AI system
can gradually enhance the efficiency and effectiveness of its reflection
process, particularly when Reflection is performed collaboratively with other AI
systems.

\section{Human Simulation Learning}

All stages of Human Simulation Computation are learnable, including thinking
modes, reflection, action execution, and activity scheduling.
These components can be continuously updated through experience.
By embedding basic thinking modes together with mechanisms for incremental
updating, the AI system can gradually mature and acquire foundational
capabilities.

Observed information is also learned to establish a representation of normal
environmental behavior.
This learned baseline can later be used for comparison, enabling the system to
detect deviations and identify salient features through difference analysis.
Learning sequential actions is particularly valuable, as it reduces future
planning effort by reusing previously learned action patterns.

\subsection{Thinking Mode for Learning}

Thinking modes themselves are learnable components of Human Simulation
Computation.
Through experience, the AI system can refine how it reasons, reflects, and
decides which strategies to apply in different situations.
Learning thinking modes allows frequently effective reasoning patterns to become
more intuitive over time, reducing computational overhead while improving
decision quality.

\subsubsection{Whole Simulation Process Learning on Time}

The learning process does not focus solely on linguistic content, but also
incorporates thinking, reflection, activity scheduling, actions, and
environmental feedback over time.
These elements collectively form the complete human simulation process, as
described in earlier sections.

To support whole-process learning, all processed or sensed information is
recorded as it occurs.
This requires two parallel mechanisms: one responsible for executing the
current task, and another for recording relevant information for learning.
The objective is to enable comprehensive and timely learning of the entire
process, rather than isolated outcomes.

\subsubsection{Max-Entropy Learning}

Whole-process on-time learning establishes a baseline of normal behavior.
Once similar patterns have been sufficiently learned, repeatedly learning them
provides diminishing value.
As in human learning, the system should prioritize acquiring new or informative
knowledge rather than reinforcing already familiar content.

In real-world environments, numerous events occur continuously, and not all of
them are equally valuable to learn.
We adopt the principle that actions induce changes in the environment~\cite{su2026actively};
therefore, events associated with noticeable changes are more likely to reflect
meaningful causal relationships.
Accordingly, learning prioritizes experiences with higher entropy, as they
contain more information.

Newly encountered issues typically exhibit higher entropy and are therefore
given priority in learning.
By focusing on high-entropy experiences rather than all observations, the system
can allocate learning resources more efficiently and improve its ability to
identify significant patterns.

\subsubsection{Broad-Scope Learning}

Human Simulation Computation is not designed around narrowly predefined tasks.
Instead, its overarching objective is to adapt and operate effectively within a
dynamic environment.
Accordingly, learning should not be restricted to specific tasks, but should
cover a broad range of situations that may arise across time, space, and
interaction with other AI systems or environmental conditions.

Because environmental states are sensed through limited and potentially noisy
channels, the AI system must learn to generalize across incomplete observations.
Broad-scope learning enables the system to acquire fundamental adaptive
capabilities that support future task-specific learning.
This form of learning establishes a foundational understanding of the
environment, serving as a basis for more specialized behaviors.

\subsubsection{Learning Sequential Actions and Their Manifestations}

Sequences of actions should also be learned.
Many human activities consist of habitual action sequences that no longer
require explicit reasoning once they are familiar.
For example, when planning to eat, washing hands is performed automatically
without repeatedly reasoning about its health benefits.
Learning such action sequences reduces future planning and computation costs.

In addition to learning action sequences, the system should also learn the
manifestations of actions.
This includes fine-grained execution details, such as motion patterns or timing,
analogous to learning how to dance.
By learning both the structure and appearance of actions, the system can execute
them more fluently and reliably.

\textbf{Checking understanding through action transfer.}
Understanding a process can be verified by the ability to translate it into
action.
If the AI system can decompose a task into executable actions and carry them out
internally or externally, it demonstrates genuine understanding.
This conversion from knowledge to action provides a practical criterion for
evaluating comprehension.

\subsubsection{Learning with Reasons}

Learning should not be limited to outcomes or procedures alone, but should also
capture the underlying reasons behind observations and decisions.
Understanding why an event occurs or why a particular method is effective
supports deeper comprehension and improves generalization.

By learning reasons in addition to facts, the AI system can generate actions to
verify hypotheses or validate explanations.
Accordingly, the learning process should explicitly encourage the system to
record causal explanations and rationales alongside observed data.
This approach strengthens the system’s ability to reason, adapt, and transfer
knowledge to new situations.

\subsection{Actions to Promote Learning}

Actions are the means by which the AI system actively changes both its external
environment and its internal processes in pursuit of its goals.
These changes affect not only task execution, but also the thinking, reflection,
and learning components of Human Simulation Computation.
Accordingly, actions play a central role in promoting effective learning.
\subsubsection{Actions for On-Time Learning}

On-time learning refers to learning that occurs concurrently with ongoing
activities.
Rather than relying on offline intervention by programmers or human operators,
the AI system should take actions to ensure that newly acquired information is
immediately incorporated into its learning process.

When an issue is new and only limited data are available, the system may perform
actions to generate additional learning material.
Such actions include actively searching for related information, probing the
environment, or independently varying certain conditions.
By expanding the available data through action, the system supports more
effective and timely learning.

\subsubsection{On-Time Learning Embedded into Other Human Simulation Processes}

The actions and thinking processes at each stage of Human Simulation Computation
should be learned as they occur.
This reflects a common human behavior: nearly everything a person does is, to
some extent, remembered and contributes to future behavior.
Accordingly, the procedures used to handle tasks should also be recorded and
learned.

For example, the outcomes of Reflection processes may be stored either through
explicit recording or by updating model parameters.
When associations are discovered during reflection, the AI system may take
additional actions to reinforce memory, such as extracting key features from
multiple perspectives to generate training data.
This is particularly important when the issue is novel or exhibits high
entropy~\cite{su2025human}.
The learning process itself should be guided by thinking, prioritizing features
that are not directly causal but frequently co-occur and carry high
informational value.
\subsubsection{Learning Normal Environmental Outcomes}

The AI system should learn the normal outcomes and patterns of its environment
based on accumulated observations.
These learned norms serve as reference baselines for future comparison, enabling
the system to detect anomalies, deviations, or potential risks.

Both internal processes and external interactions contribute to this learning.
Internal reasoning states, action execution results, sensor data from the
environment, and feedback from other AI systems are all recorded and integrated.
By learning these normal patterns on time, the system improves its ability to
identify significant differences and respond appropriately in subsequent
situations.
\subsubsection{Improving Learning Through Thinking and Reflection}
\label{sec_active_learn}

Thinking and Reflection processes play an important role in refining the learning
mechanism itself.
Through reflection, the system can evaluate whether current learning strategies
are effective or whether they lead to suboptimal outcomes in later actions.

Thinking may propose improved ways to prepare training materials, select
representative data, or determine which information should be explicitly
recorded.
Reflection can identify learning methods that resulted in failure or inefficiency
and guide adjustments to future learning strategies.
By continuously refining learning through thinking and Reflection, the AI system
achieves more efficient and adaptive knowledge acquisition.

\section{Action to Interact with the Environment, Other AI Objects, and Task Execution}

Actions can be applied not only to the internal processes of Human Simulation
Computation, but also to actively interact with the external environment and
other AI systems~\cite{su2026actively}.
Such interactions aim to modify external conditions in order to improve the
system’s ability to live and operate effectively within its environment.

Actions serve multiple roles.
They enable participation in the environment, respond to feedback or requests
from other agents, and execute tasks when appropriate.
Whether to accept a task from another agent and how to perform it are decisions
made by the AI system based on its own thinking and goals.

\subsection{Common Thinking Strategies for Action}

Before executing actions, the AI system applies common thinking strategies to
evaluate priorities, potential consequences, and long-term impact.
These strategies guide action selection to ensure that behavior aligns with the
system’s goals and constraints, rather than being purely reactive.

\subsubsection{Basic Requirements for Survival}

Actions that ensure the basic ability of the AI system to operate are given the
highest priority.
These include maintaining sufficient power supply, preventing physical or
functional damage, and preserving system stability.
Such actions constitute the first level of scheduling and must be satisfied
before pursuing higher-level objectives.

Once basic requirements are secured, the system can allocate resources to
actions that improve performance, comfort, or long-term benefits.
This hierarchy mirrors human behavior, where survival and stability take
precedence over optimization or exploration.

\subsubsection{Action with Long-Term Benefit Consideration}

Benefits may manifest over different time horizons.
Some actions yield immediate gains, while others provide advantages only in the
long term.
Accordingly, the AI system should evaluate and compare both short-term and
long-term benefits when selecting actions.

For example, driving at high speed may reduce travel time but increase the risk
of damage or discomfort.
Such trade-offs should be explicitly considered and may be provided to the LLM
for further evaluation.
Feedback from the environment, other AI systems, and the system itself should be
integrated into this assessment~\cite{su2026actively}.

Interactions with other AI systems are particularly important, as their feedback
often reflects external constraints or shared objectives.
While each agent ultimately prioritizes its own goals, decisions should at
least ensure basic survival and avoid unnecessary harm to others.

\subsubsection{Verifying Candidate Methods in the Environment}

When multiple candidate methods are available, the AI system may rely on
environmental feedback to determine which option is most effective.
Such feedback provides an objective basis for evaluating whether internal
reasoning and predictions align with real-world outcomes.

Before executing an action, the system may plan several alternative solutions
and select the most promising one based on expected results.
If an action causes discomfort or negative internal states, the system may
search for alternative methods that reduce adverse effects.
By verifying candidate methods through interaction with the environment, the
system continuously improves the accuracy of its decision making.

\subsubsection{Learning Feedback from the Environment}

Feedback obtained from the environment should be treated as an important source
of learning.
Environmental responses provide objective signals that help the AI system assess
whether its internal reasoning and chosen actions are appropriate.

All sensed feedback is processed by the system and simultaneously passed to the
learning component.
Over time, this feedback contributes to forming expectations about action
outcomes and supports more accurate prediction and decision making in future
interactions.

\subsection{Actions to Optimize Interactive Behavior}

To function effectively in dynamic environments, the AI system must continuously
optimize how it interacts with external entities.
These interactions are not static; they evolve in response to changes in time,
location, environmental conditions, and feedback from other AI systems.
Accordingly, action strategies should be adaptable and guided by ongoing
thinking and evaluation.

\subsubsection{Dynamic Interaction Actions Guided by Thinking}

Actions used to interact with the environment must be dynamically adjusted, as
real-world conditions change frequently across time, location, and interaction
context.
Such changes may arise from environmental variation, movement across locations,
or feedback from other AI systems.

Thinking guides the selection and adaptation of interaction actions.
For example, when an AI system transitions from one road segment to another, the
new environment may introduce unfamiliar constraints or risks.
This change triggers renewed thinking to assess the situation and determine
appropriate actions.
By continuously integrating thinking with interaction, the system maintains
flexibility and responsiveness in complex environments.

\subsubsection{Sequential Actions for Learning and Reflection}

\textbf{Sequential Actions for Learning.}
Certain actions frequently occur together and can be learned as fixed sequences.
Once such sequences are familiar, the system no longer needs to plan each step
explicitly.
This behavior mirrors human habits, where familiar routines are executed with
minimal deliberation, thereby reducing computational overhead.

When environmental conditions change, these sequences can be reorganized and
relearned as needed.
This balance between habitual execution and adaptive restructuring supports
efficient yet flexible behavior.

\textbf{Reflection}
After action execution, the system performs Reflection to evaluate outcomes and
identify potential improvements.
By reflecting on the results of sequential actions, the system can refine both
the action sequence and the underlying thinking strategies to achieve better
performance in future interactions.

\subsubsection{Obtaining Comprehensive Feedback}

Feedback from the environment should be collected as comprehensively as possible.
This includes internal outcomes, external environmental responses, and feedback
from other AI systems.
Comprehensive feedback provides a reliable basis for evaluating action quality
and guiding subsequent thinking and learning.

Sensor data with relatively high accuracy should be recorded to measure feedback
intensity and detect potential risks.
This includes explicit error signals, abnormal sensor readings, or values that
approach dangerous thresholds within short time intervals.
In addition to numerical measurements, qualitative feedback can also be assessed
through thinking or prediction mechanisms, such as prompting an LLM to estimate
severity or impact.

Internal system states should also be treated as feedback.
Recording internal predictions and reasoning results enables comparison with
actual outcomes, helping the system identify discrepancies and refine future
decisions.

\subsubsection{Using Sufficient Action Channels for Interaction}

Humans use multiple action channels—such as body movement, facial expression,
and language—to convey intentions, emotions, and requests.
Similarly, a human simulation system should be equipped with sufficiently rich
action execution mechanisms to interact effectively with other AI systems and
the environment.

Providing diverse action channels allows the AI system to express nuanced
responses, transmit feedback, and coordinate with other agents more precisely.
Such expressiveness enhances interaction quality and supports more robust and
human-like collaboration.

\section{Proof}

In this section, we provide a brief theoretical justification for Human
Simulation Computation (HSC).
Specifically, we argue that (1) active interaction with the environment is
necessary to improve and verify the human simulation process, and (2) human
simulation strategies cannot be fully learned from language data alone.
Together, these arguments demonstrate the necessity of incorporating
human-inspired simulation mechanisms beyond purely language-based learning.

\subsection{Improvement and Verification Through Environmental Interaction}

To live better in an environment, an AI system must execute the human simulation
loop \emph{within} the environment rather than reasoning in isolation.
Environmental interaction provides two essential benefits: (i) it improves the
agent by generating new experience, and (ii) it verifies whether the agent’s
internal reasoning is consistent with reality.

Let $s_t$ denote the internal cognitive state at time $t$, and let $f_t$ denote
environmental factors observed at $t$.
Under HSC, the agent produces an action $a_t$ and receives environmental
feedback $o_{t+1}$ (observation) and possibly a scalar utility signal $u_{t+1}$
that reflects ``living better'' (e.g., safety, resource stability, goal
satisfaction).
A minimal interaction model can be written as:
\begin{equation}
\label{eq:env_transition}
o_{t+1} \sim P(\cdot \mid o_t, a_t), 
\qquad 
u_{t+1} = U(o_t, a_t, o_{t+1}),
\end{equation}
and the internal state update follows a learning/reflection operator:
\begin{equation}
\label{eq:hsc_update_proof}
s_{t+1} = \mathcal{L}\!\left(s_t,\; \mathcal{R}(s_t, a_t, o_{t+1}, u_{t+1})\right).
\end{equation}

\textbf{Verification effect.}
Let $\hat{o}_{t+1} = g(s_t, a_t)$ be the agent’s predicted consequence of taking
$a_t$ under its internal model.
A simple discrepancy (verification) signal is:
\begin{equation}
\label{eq:verification_error}
e_{t+1} = d\!\left(o_{t+1}, \hat{o}_{t+1}\right),
\end{equation}
where $d(\cdot,\cdot)$ is a nonnegative distance (e.g., $\ell_2$ for numeric
signals or $1-\text{sim}$ for semantic similarity).
If the agent does not interact with the environment, then $o_{t+1}$ is not
observed and $e_{t+1}$ cannot be computed, so incorrect internal beliefs may
persist.
With interaction, $e_{t+1}$ provides a grounded criterion to validate or reject
internal reasoning and to guide $\mathcal{R}$ and $\mathcal{L}$.

\textbf{Improvement effect.}
Assume the learning operator reduces the expected discrepancy, i.e.,
\begin{equation}
\label{eq:error_decrease}
\mathbb{E}\!\left[e_{t+1} \mid s_{t+1}\right] \le 
\mathbb{E}\!\left[e_{t+1} \mid s_t\right],
\end{equation}
which captures the standard notion that feedback-driven updates improve the
internal model.
Then repeated interaction produces a sequence of states $\{s_t\}$ with improved
predictive consistency.
Moreover, since actions are executed step by step, the agent can obtain earlier
and richer feedback than a single final outcome, enabling finer-grained
correction of thinking, reflection, and learning.

Overall, environment interaction expands the available information scope and
provides objective signals to verify and refine internal reasoning, which is
necessary for HSC to achieve adaptive ``living better'' behavior.

\subsection{Human Simulation Strategies Cannot Be Fully Learned from Language Material}

Although large language models acquire extensive knowledge from textual data,
human simulation strategies cannot be fully derived from language material
alone.
Language reflects the outcomes and descriptions of human thinking, but it does
not provide complete access to the embodied, interactive processes through
which thinking is validated and refined in real environments.

\subsubsection{Unlimited Learning from the Environment Through Action}

The environment is dynamic and open-ended, and its possible states cannot be
fully enumerated or guaranteed in advance.
Accordingly, the space of conditions encountered by an AI system through
interaction is effectively unbounded.
Actions taken under such conditions may vary widely and generate outcomes that
are not present in static language corpora.

Formally, let $\mathcal{E}$ denote the environment state space and $\mathcal{A}$
the action space.
Language data provide samples from a descriptive distribution
$P_{\text{text}}(x)$ over symbolic expressions.
In contrast, interaction generates samples from a transition distribution
$P_{\text{env}}(e_{t+1} \mid e_t, a_t)$, where:
\begin{equation}
\label{eq:env_open}
|\mathcal{E}| \rightarrow \infty, 
\qquad 
|\mathcal{A}| \rightarrow \infty,
\end{equation}
in realistic settings.

Because human behavior is not constrained to predefined measurements or fixed
action sets, humans can adapt to unbounded environments.
Actions provide a mechanism to probe, verify, and refine understanding under
previously unseen conditions.
Through action, the AI system can test hypotheses, validate assumptions, and
establish causal relations that cannot be inferred from language alone.

Moreover, actions promote internal reasoning processes, including thinking,
learning, and reflection, by grounding abstract knowledge in concrete outcomes.
This grounding enables self-growth and adaptation across a broader range of
scenarios than language-only learning permits.

\subsubsection{Dynamic Processes and Self-Growth}

All components of Human Simulation Computation are inherently dynamic.
Thinking, action, learning, and reflection continuously influence one another
and are updated through experience.
This dynamic coupling enables self-growth that cannot be achieved through
static knowledge acquisition alone.

Let $\Theta_t$ represent the set of internal strategies used by the agent at time
$t$, including thinking modes, action policies, and learning rules.
Under HSC, these strategies evolve according to:
\begin{equation}
\label{eq:strategy_update}
\Theta_{t+1} = \Theta_t + \Delta\Theta(s_t, a_t, o_{t+1}),
\end{equation}
where $\Delta\Theta(\cdot)$ is induced by learning and reflection.
Without interaction, $\Delta\Theta$ is driven only by symbolic consistency,
whereas interaction introduces grounded correction signals.

Because actions can actively change the environment, the agent can intentionally
create informative situations rather than waiting for them to occur.
This capability allows progress toward specific purposes, such as improving
robustness or achieving long-term goals.
Stage-wise thinking and reflection guide each update, ensuring that growth is
directed rather than random.

\subsubsection{Rationality of Human Thinking Strategies}

Human thinking strategies are rational in the sense that they have been shaped,
validated, and refined through long-term interaction with real environments.
These strategies are not arbitrary heuristics, but outcome-oriented reasoning
patterns that have repeatedly demonstrated effectiveness in achieving survival,
adaptation, and goal satisfaction.

From a decision-theoretic perspective, human thinking strategies can be viewed
as approximate solutions to sequential decision problems under uncertainty.
Let $\pi_h$ denote a human-inspired thinking policy and let $J(\pi)$ be a utility
measure associated with ``living better'' in the environment.
Through continuous interaction and feedback, human strategies implicitly seek
to maximize expected utility:
\begin{equation}
\label{eq:rational_thinking}
\pi_h \approx \arg\max_{\pi} \; \mathbb{E}\!\left[ \sum_{t=0}^{T} \gamma^t
u_t \,\middle|\, \pi \right],
\end{equation}
where $u_t$ is the utility signal derived from environmental feedback and
$\gamma \in (0,1]$ is a discount factor.

Language data encode descriptions of these strategies, but not the full feedback
loops through which they were validated.
By explicitly embedding human-inspired thinking strategies into HSC and allowing
them to be continuously tested and updated through interaction, the AI system
inherits not only symbolic rationality but also empirically grounded rational
behavior.

\subsection{Reducing the Number of Candidate Considerations by Difference}

Human-inspired reasoning can be viewed as a class of heuristic algorithms whose
primary advantage lies in reducing the number of candidate solutions that must
be considered.
Instead of exhaustively enumerating all possible actions or reasoning paths, the
system focuses on differences that distinguish the current situation from
normal or expected conditions.

Let $\mathcal{C}$ denote the set of all candidate actions or reasoning paths.
Without difference-based filtering, the system must evaluate $|\mathcal{C}|$
candidates, which may be prohibitively large.
By introducing a difference function $\Delta(\cdot)$ that measures deviation
from a baseline or expected state, the candidate set can be reduced to:
\begin{equation}
\label{eq:diff_filter}
\mathcal{C}' = \{ c \in \mathcal{C} \mid \Delta(c) > \tau \},
\end{equation}
where $\tau$ is a threshold that controls sensitivity to differences.

Candidates with larger differences are more likely to be informative or
critical.
As the magnitude of difference increases, the number of viable candidates
decreases, allowing the system to concentrate computational resources on a
smaller, more relevant subset.
This mechanism explains why difference-based reasoning can significantly reduce
search complexity while maintaining effectiveness.

Therefore, by prioritizing differences, Human Simulation Computation achieves
efficient exploration and decision making.
This property further supports the necessity of human-inspired strategies in
complex, open-ended environments where exhaustive consideration is infeasible.

\section*{Conclusion}

In this paper, we proposed \emph{Human Simulation Computation} (HSC), a systematic
framework that enables AI systems to adapt to complex and dynamic environments
by simulating essential human cognitive and behavioral processes.
Unlike conventional task-oriented AI systems or language-model–centric methods,
HSC emphasizes continuous interaction with the environment, treating thinking,
learning, reflection, action, and feedback acquisition as tightly coupled and
mutually reinforcing stages.

We argued that language-based learning alone is insufficient for acquiring
human-like adaptability.
By explicitly introducing action as an active mechanism—not only for task
execution but also for verification, information acquisition, and learning
promotion—HSC allows AI systems to ground internal reasoning in environmental
feedback.
Thinking guides decision-making across stages, reflection enables self-correction
and method refinement, and on-time learning accumulates experience from both
internal processes and external interactions.
Together, these components form a closed-loop system oriented toward long-term
survival and improvement rather than isolated task completion.

In addition, we summarized commonly used human thinking strategies and showed how
they can be embedded into different stages of the HSC process to reduce search
space, improve decision quality, and enhance robustness under uncertainty.
By incorporating scheduling mechanisms and activity control, HSC further supports
autonomous operation during both active and idle periods, enabling continuous
self-improvement without constant external supervision.

Future work will focus on implementing HSC in practical AI systems, including
embodied agents and multi-agent environments, and on studying efficient mechanisms
for integrating action-driven feedback with large language models.
Another important direction is to formally analyze the convergence and stability
properties of long-term human simulation processes.
We believe that HSC provides a promising foundation for developing AI systems that
not only solve tasks, but also \emph{live, adapt, and grow} within their
environments in a human-inspired manner.


\ifCLASSOPTIONcaptionsoff
  \newpage
\fi

\bibliographystyle{IEEEtran}
\bibliography{ref}

%

\begin{IEEEbiography}{Hong Su}
  received the MS and PhD degrees, in 2006 and 2022, respectively, from Sichuan University, Chengdu, China. He is currently a researcher of Chengdu University of Information Technology Chengdu, China. His research interests include blockchain, cross-chain and smart contract.
\end{IEEEbiography}




\end{document}